\documentclass[letterpaper]{article}
\usepackage{aaai2026} 
\usepackage{times} 
\usepackage{helvet} 
\usepackage{courier} 
\usepackage[hyphens]{url} 
\usepackage{graphicx} 
\usepackage{graphicx} 
\usepackage{natbib} 
\usepackage{caption} 
\usepackage{algorithm}
\usepackage{algorithmic}
\usepackage{newfloat}
\usepackage{listings}
\DeclareCaptionStyle{ruled}{labelfont=normalfont,labelsep=colon,strut=off} 
\floatstyle{ruled}
\newfloat{listing}{tb}{lst}{}
\floatname{listing}{Listing}

\usepackage{amsmath}
\usepackage{cleveref}
\usepackage{amssymb}
\usepackage{longtable}
\usepackage{booktabs} 
\usepackage{array}    
\usepackage{tabularx} 
\usepackage{makecell} 
\usepackage{multirow} 
\usepackage[table]{xcolor} 
\title{MAUGen: A Unified Diffusion Approach for Multi-Identity Facial Expression and AU Label Generation}
\author {
    Xiangdong Li\textsuperscript{\rm 1},
    Ye Lou\textsuperscript{\rm 1},
    Ao Gao\textsuperscript{\rm 1},
    Wei Zhang\textsuperscript{\rm 1,2}\thanks{Corresponding authors},
    Siyang Song\textsuperscript{\rm 3*}
}
\affiliations {
    \textsuperscript{\rm 1}School of Software Technology, Zhejiang University\\
    \textsuperscript{\rm 2}Innovation Center of Yangtze River Delta, Zhejiang University\\
    \textsuperscript{\rm 3}HBUG Lab, Department of Computer Science, University of Exeter\\ 
    \{xiangdong.li, ye\_lou, gaoao.olivia, cstzhangwei\}@zju.edu.cn, s.song@exeter.ac.uk\\
}

\begin{document}
\maketitle

\begin{abstract}

The lack of large-scale, demographically diverse face images with precise Action Unit (AU) occurrence and intensity annotations has long been recognized as a fundamental bottleneck in developing generalizable AU recognition systems. In this paper, we propose MAUGen, a diffusion-based multi-modal framework that jointly generates a large collection of photorealistic facial expressions and anatomically consistent AU labels, including both occurrence and intensity, conditioned on a single descriptive text prompt. Our MAUGen involves two key modules: (1) a Multi-modal Representation Learning (MRL) module that captures the relationships among the paired textual description, facial identity, expression image, and AU activations within a unified latent space; and (2) a Diffusion-based Image-label Generator (DIG) that decodes the joint representation into aligned facial image-label pairs across diverse identities. Under this framework, we introduce Multi-Identity Facial Action (MIFA), a large-scale multi-modal synthetic dataset featuring comprehensive AU annotations and identity variations. Extensive experiments demonstrate that MAUGen outperforms existing methods in synthesizing photorealistic, demographically diverse facial images along with semantically aligned AU labels.
\end{abstract}
\begin{links}
\link{Code}{https://github.com/XDLI13/MAUGen/tree/main}
\end{links}

\section{Introduction}
\label{sec:intro}

\noindent Human Facial Action Units (AUs), defined by the Facial Action Coding System (FACS)~\cite{ekman1978facial}, encode muscle movements underlying facial expressions. They have been widely used in affective computing~\cite{song2022spectral}, human-computer interaction~\cite{song2025react}, and psychology~\cite{donato1999classifying} for quantifying emotional behaviors. Although reliable AU recognition models are essential for developing various real-world intelligent systems, their performance generalization, however, heavily depends on the quality, scale, and diversity of paired face–AU training data.

\begin{figure}[th!]
    \centering
    \includegraphics[width=0.47\textwidth]{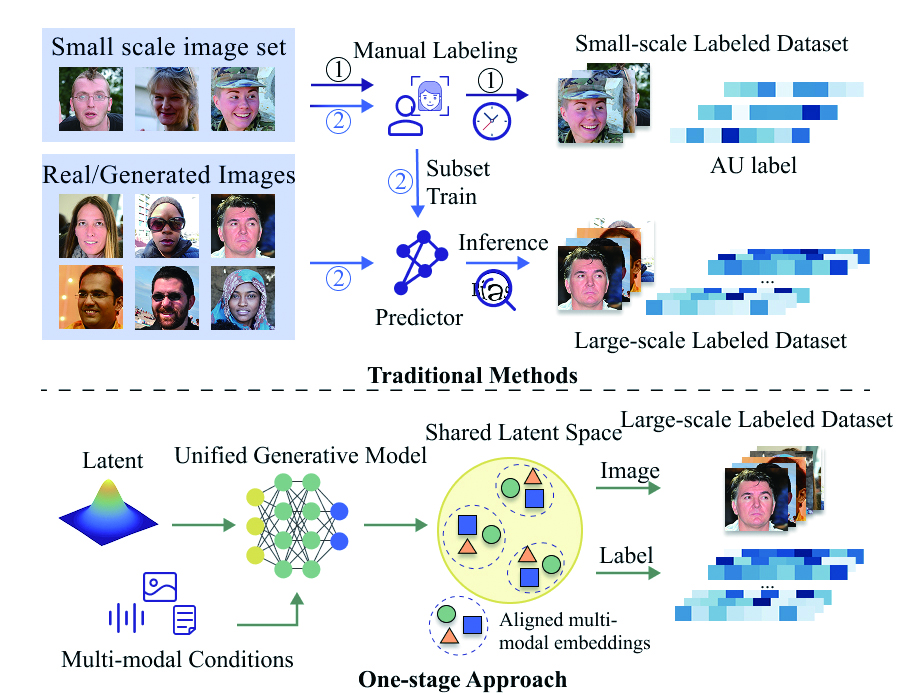}
    \caption{
       Comparison of two traditional AU-labeled dataset construction pipelines (indexed numerically) with our unified one-stage approach.}
    \label{fig:strategy}
\end{figure}

\begin{figure*}[th!]
    \centering
    \includegraphics[width=1\textwidth]{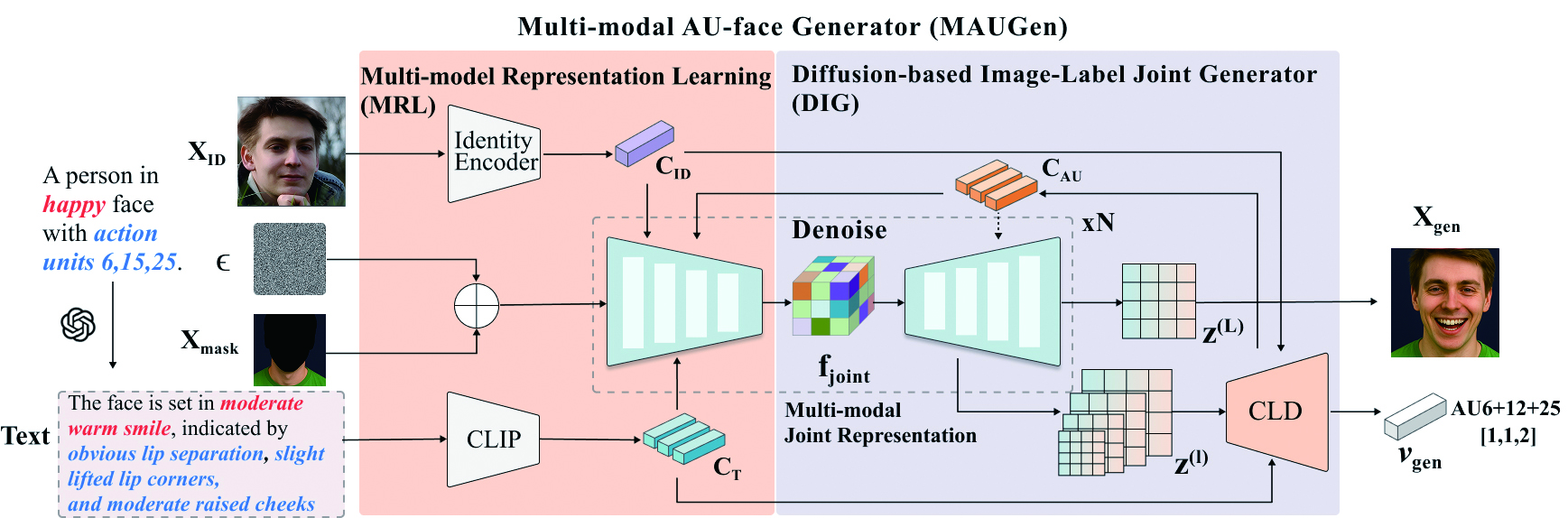} 
    \caption{
        Overview of our MAUGen framework. The user-defined prompt is first expanded into detailed expression descriptions using an LLM. The resulting text features, together with identity exemplars, facial masks, and structure-aware AU embeddings, are encoded into a joint latent space by the MRL module (\Cref{sec:mre}). The DIG module (\Cref{sec:DIG}) subsequently synthesizes identity-consistent facial images and corresponding AU labels via an image decoder and the Conditional Label Decoder (CLD).
    }
    \label{fig:proposed_method}
\end{figure*}

While recent advances in deep learning have significantly advanced facial Action Unit (AU) recognition, its progress remains hindered by the limitations of available datasets. Manual AU annotation is resource-intensive and constrained by privacy and FACS expert requirements, limiting its scalability. Consequently, widely used benchmarks such as DISFA~\cite{mavadati2013disfa}, BP4D~\cite{zhang2014bp4d}, and CK+~\cite{lucey2010extended} contain a limited number of identities, exhibit excessive frame redundancy (e.g., near-duplicate video frames)~\cite{hu2022facial}, and often lack AU intensity annotations~\cite{kollias2019expression}, thereby limiting their applicability for robust and fine-grained AU modeling. Large-scale in-the-wild datasets such as AffectNet~\cite{mollahosseini2017affectnet}, EmotionNet~\cite{fabian2016emotionet}, and FABA-Instruct~\cite{li2025facial} improve diversity but often suffer from weak or inconsistent labels. These limitations stem from automatic annotations using pre-trained DL models or LLMs, which are prone to errors under uncontrolled backgrounds and lack rigorous validation with pipelines shown in~\Cref{fig:strategy}. For instance, only 5\% of the data in EmotionNet is used for the predictor training, raising the concern about the label reliability. 

To address this challenge, we aim to automatically generate a large-scale, well-annotated, and demographically diverse multi-modal AU dataset. Earlier image synthesis efforts have leveraged pretrained Generative Adversarial Networks (GANs) to synthesize images with semantic masks, where predictors such as MLPs~\cite{li2022bigdatasetgan} or GAN inversion~\cite{xu2023handsoff} enable the automatic annotation of large-scale unlabeled images by training on small subsets. However, such methods still rely on partially labeled data, where the trained predictors are prone to systematic errors/bias, particularly due to artifacts introduced by GAN inversion~\cite{xia2022gan}. Alternatively, diffusion-based image generation models such as GLIDE~\cite{nichol2022glidephotorealisticimagegeneration}, DALL-E 2~\cite{ramesh2022hierarchical}, Imagen~\cite{saharia2022photorealistic}, and Stable Diffusion~\cite{rombach2022high}, which have unified multi-modal pipelines and adopt joint latent learning, enable one-stage co-generation of images and labels.
These diffusion-based co-synthesis frameworks~\cite{yu2023diffusion} further promote scalable and automatic dataset generation. This motivates the design of a unified one-stage pipeline for multi-modal facial behavior synthesis (see \Cref{fig:strategy}). Nonetheless, achieving fine-grained control and high visual fidelity remains challenging, due to the ambiguity in detailed textual descriptions~\cite{song2020score} and the architectural constraints~\cite{rosenberg2024limitations}.

In this paper, we propose MAUGen, a novel Multi-modal AU-face Generator that can jointly synthesize a large-scale dataset containing diverse facial expression images, identity-agnostic AU labels, and fine-grained textual descriptions from a user-defined prompt. It comprises two key modules: (i) the Multi-modal Representation Learning (MRL) module that constructs a unified latent space by integrating identity, text, and implicit AU structural semantics, employing a mutual conditioning strategy for cross-modal consistency; and (ii) the Diffusion-based Image-Label Joint Generator (DIG) that jointly decodes facial images and AU labels, enhanced by a triplet self-supervision mechanism and a language-guided optimization strategy. This way, our MAUGen enables accurate generation of facial expressions and their AU activations with highly detailed semantic control. The overall framework is shown in~\Cref{fig:proposed_method}. Our key contributions and novelties are summarized as follows:
\begin{itemize}

\item We propose \textbf{MAUGen}, a unified multi-modal generation framework which can jointly produce expression-rich facial images and identity-agnostic AU labels from a single textual prompt.

\item We propose novel cross-modal alignment strategies in both encoding and decoding stages, enabling the generation of images with diverse expressions and identities, as well as the corresponding identity-agnostic AU labels. 

\item Extensive experiments show that MAUGen achieves robust multi-modal alignment. We also release the Multi-Identity Facial Action (MIFA) dataset, which contains over one million annotated images, to advance future research in AU recognition and expression analysis.

\end{itemize}

\begin{figure*}[t]
    \centering
    \includegraphics[width=1\textwidth]{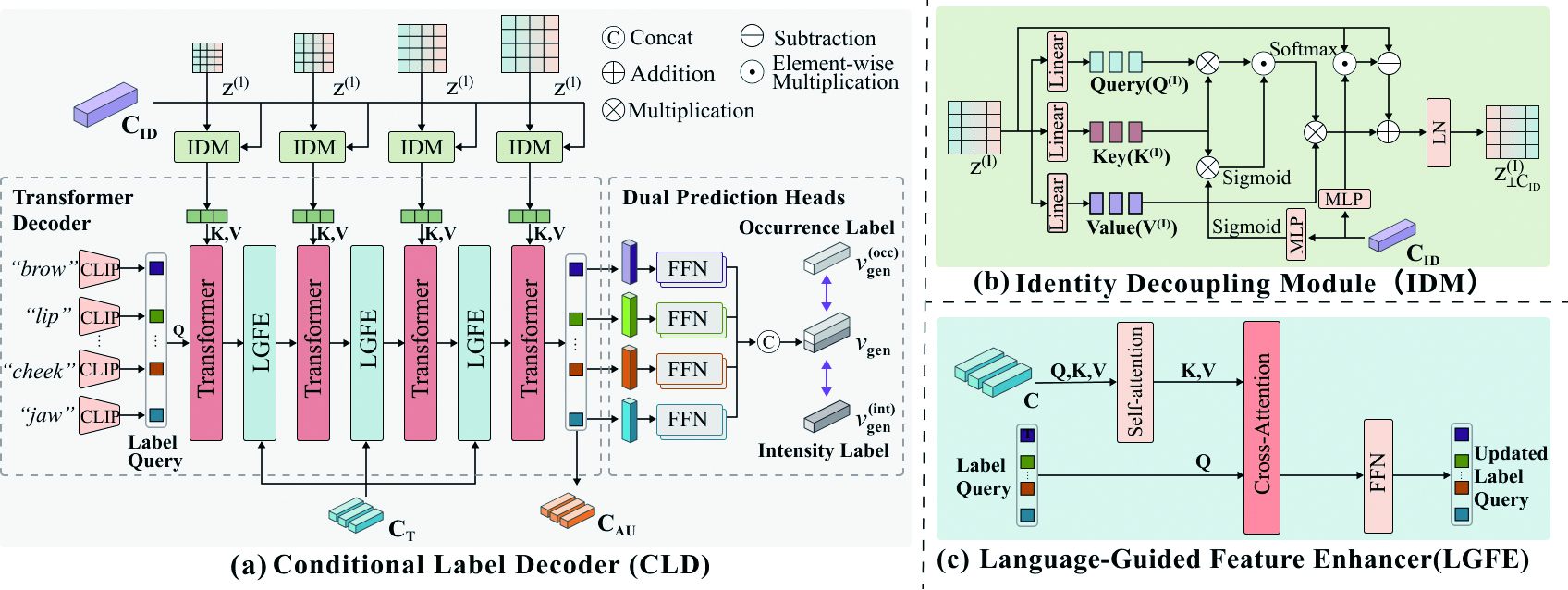}
    \caption{
        Structure of Conditional Label Decoder (CLD). (a) The AU queries initialized from text embeddings are refined via a Transformer Decoder with dual prediction heads and Identity Decoupling Modules (IDMs) to predict occurrence and intensity labels. (b) The IDM removes identity-related components from latent features with a modulation mask and residual filtering. (c) The Language-Guided Feature Enhancer (LGFE) injects the token-wise global semantic context into AU queries via attention.
    }
    \label{fig:CLD}
\end{figure*}

\section{Related Work}
\label{sec:related}

\noindent\textbf{Face image synthesis:} 
Recent advances in facial image synthesis have been primarily driven by two paradigms: generative adversarial networks (GANs) and diffusion models. GAN-based methods, such as the StyleGAN series~\cite{karras2019style}, produce high-fidelity images with structured latent spaces conducive to editing and disentanglement. Diffusion models~\cite{ho2020denoising,rombach2022high} offer improved sample diversity and robustness. To support more precise control over facial attributes, recent efforts have introduced diverse conditioning strategies. Text-guided facial manipulation has been explored across both GAN~\cite{patashnik2021styleclip,xia2021tedigan} and diffusion-based~\cite{sun2022anyface,sun2024cemiface} frameworks, with recent extensions incorporating multi-modal prompts for fine-grained expression control. Identity preservation remains a central topic, often addressed through contrastive objectives or reference-image guidance to retain subject-specific traits across expressions and styles~\cite{tang2024ipdm,shiohara2024face2diffusion}. Semantic priors such as facial masks and predefined AU features have also been employed to guide localized expression synthesis~\cite{he2024synferboostingfacialexpression,varanka2024localizedfinegrainedcontrolfacial}. However, existing methods still face limitations in achieving semantically consistent expression control across diverse conditions. To address this, we propose a unified multi-modal generation framework that encodes textual semantic, identity, and AU co-activation representations into a shared latent space, enabling simultaneous synthesis of photorealistic facial images and their corresponding AU labels.

\noindent\textbf{Facial Action Unit Recognition:}
Recent advancements in AU recognition are largely driven by deep learning frameworks, including AU occurrence recognition \cite{yuan2024auformer} and AU intensity recognition \cite{ntinou2021transfer,song2021self}. Modeling inter-AU relationships remains fundamental for capturing co-activation patterns~\cite{luo2022learning,wang2024multi}, while generative approaches mitigate data scarcity by synthesizing AU-aware facial dynamics~\cite{yin2024fg}. Transformer-based architectures enhance robustness via long-range dependency modeling on RGB inputs~\cite{liu2024multi}. Multi-modal strategies leverage complementary cues from depth or thermal data~\cite{zhang2024multimodal}, and joint visual-semantic embeddings to enrich AU representations~\cite{yang2021exploiting}. Meanwhile, recent learning paradigms such as contrastive learning~\cite{chang2022knowledge}, weakly supervised learning~\cite{zhang2023weakly}, semi-supervised learning~\cite{tang2021piap}, and large-scale self-supervised pretraining~\cite{ning2024representation} improve the utilization of unlabeled data. Despite these advances, AU model development remains hindered by the limited scale and diversity of existing datasets, restricting generalization to complex real-world scenarios.

\begin{figure*}[t]
    \centering
    \includegraphics[width=1\textwidth]{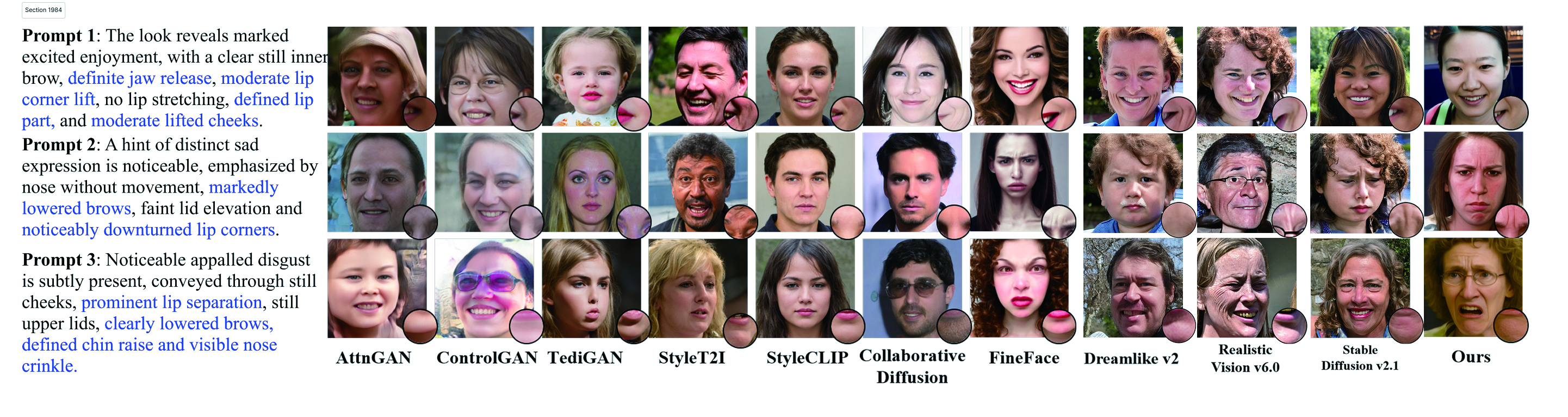} %
    \caption{
        Qualitative comparison with prior methods. MAUGen produces photorealistic facial images with enhanced semantic alignment to textual prompts, accurately capturing fine-grained expression details. Key facial regions are highlighted.}
    \label{fig:comparison}
\end{figure*}

\section{Methodology}
\label{sec:method}

\subsection{MAUGen Pipeline}
\label{sec:pipeline}

\noindent As shown in~\Cref{fig:proposed_method}, our MAUGen starts with applying a prompt-engineered large language model (LLM) to generate a set of diverse textual descriptions from a single textual prompt, yielding a set of facial expression descriptions detailing the activated AUs with human verification. The intensity of the AU label $\smash{v_{\text{pre}}^{(int)}}$ is also incorporated in the description generation process at the training stage (details and examples are provided in \textit{Appendix A.1}). Building on these generated descriptions, the \textbf{Multi-modal Representation Learning (MRL)} module encodes expressive semantics by integrating the expression prompt $\smash{T^{(m)}}$, an identity exemplar $ \smash{x_\text{ID}^{(n)}} $ randomly sampled from an identity bank, a user-defined facial mask $ x_\text{mask} $ for background control, structure-aware AU cues $ \smash{C_\text{AU}^{(m)}} $ inferred during label generation (will be discussed in~\Cref{sec:DIG}), and a noisy latent variable $ \epsilon $ into a joint latent facial representation $\smash{f_\text{joint}^{(m,n)}}$ as:
\begin{equation}
\label{eq:MRL}
    f_\text{joint}^{(m,n)} = \text{MRL}(\epsilon,x_\text{mask}, x_\text{ID}^{(n)},T^{(m)}, C_\text{AU}^{(m)})
\end{equation}
The obtained $\smash{f_\text{joint}^{(m,n)}}$ facilitates semantically aligned expression information, and serves as the input to the \textbf{Diffusion-based Image-label Joint Generator (DIG)} to further generate the facial expression image $ \smash{x_\text{gen}^{(m,n)}} $ and its corresponding identity-agnostic AU labels $ \smash{v_\text{gen}^{(m)}} $ as:
\begin{equation}
    {x_\text{gen}^{(m,n)}}, v_\text{gen}^{(m)} = \text{DIG}(f_\text{joint}^{(m,n)})
\end{equation}
This way, the proposed MRL and DIG modules together form an end-to-end encoder–decoder backbone, enabling the joint generation of expression-rich facial images and textually consistent AU labels.

\noindent \textbf{MIFA Construction:} Based on the proposed MAUGen, a large-scale \textbf{Multi-modal Facial AU (MIFA)} dataset is established. The MIFA contains over a million synthesized but photorealistic and demographically diverse facial images, with each paired with 12 AU occurrence labels, 12 6-scale AU intently labels and a textual description. To guarantee the quality and reliability of our MIFA dataset, samples are filtered based on label–prompt consistency using Jaccard similarity, along with an image realism discriminator. The dataset includes 12 balanced AU labels spanning 7 primary emotions, 6 major ethnic groups, and ages ranging from children to the elderly. More details, statistics and the dataset construction pipeline are provided in \textit{Appendix B.2–B.3}.

\begin{table}[t]
\centering
\renewcommand{\arraystretch}{1.3}
\fontsize{9pt}{9pt}\selectfont
\setlength{\tabcolsep}{2pt} 
\begin{tabular}{lcc|cc}
\toprule
 &  \multicolumn{2}{c}{DISFA} &  \multicolumn{2}{c}{BP4D} \\
    \textbf{Models}    & \textbf{FID} & \textbf{CLIP} & \textbf{FID} & \textbf{CLIP}  \\
\hline
\makecell[l]{AttnGAN~\cite{xu2018attngan}}          & 8.288 & 0.237   & 9.204 & 0.252 \\
\hline
\makecell[l]{ControlGAN~\cite{li2019controllable}}   & 7.756 & 0.233   & 9.619 & 0.258  \\
\hline
\makecell[l]{TediGAN~\cite{xia2021tedigan}}           & 7.054 & 0.251 & 7.321 & 0.286  \\
\hline 
\makecell[l]{StyleCLIP~\cite{patashnik2021styleclip}}    & 6.899 & \textbf{0.303}  & 6.803 & \underline{0.313} \\
\hline
\makecell[l]{StyleT2I~\cite{li2022stylet2i}}           & 7.550 & 0.241  & \underline{6.563} &0.268  \\
\hline
\makecell[l]{Collab Diff~\cite{huang2023collaborative} } & \underline{5.949} & 0.238 & 7.992 &0.285  \\
\hline
\makecell[l]{FineFace~\cite{varanka2024localizedfinegrainedcontrolfacial} } & 6.754 & 0.262 & 7.156 & 0.243   \\
\hline
\makecell[l]{Dreamlike 2~\cite{dreamlikev2}} & 6.919 & 0.231 & 6.955 & 0.277   \\
\hline
\makecell[l]{Realistic Vision V6~\cite{realisticvisionv6}} & 6.497 & 0.228 & 6.668 & 0.276   \\
\hline
\makecell[l]{SD 2~\cite{rombach2022high} } & 5.741 & 0.229 & 6.659 & 0.275   \\
\hline
\textbf{MAUGen (Ours)} & \textbf{5.516} & \underline{0.295}  & \textbf{6.517} & \textbf{0.320} \\
\bottomrule

\end{tabular}
\caption{ \fontsize{10pt}{10pt}\selectfont Quantitative results on DISFA and BP4D. Models are fine-tuned per dataset. The best and second-best scores are in bold and underline, respectively. }
\label{tab:quantitative_comparisons}
\end{table}

\subsection{Multi-modal Representation Learning}
\label{sec:mre}

\noindent As shown in~\Cref{fig:proposed_method}, the proposed MRL module extends the downsampling path of the \textit{denoising UNet}~\cite{ho2020denoising} to incorporate multi-modal conditions, as the encoding stages of conditional diffusion models provide a stable and information-rich domain for cross-modal feature alignment. It aims to address the problem that frequently suffered by previous diffusion-based face generation models, i.e., limited semantic consistency when handling long or ambiguous textual prompts, particularly with AU details due to the absence of explicit structural priors and the lack of an effective mechanism to align AU semantics with visual representations \cite{shi2025expertgentrainingfreeexpertguidance}. 
Specifically, we propose a \textbf{Multi-modal Mutual Conditioning} strategy which introduces a series of AU co-activation representations $\smash{C_\text{AU}^{(m)}(t)}$ ($t = 0, 1, \cdots, T$) as auxiliary semantic priors to constrain the denoising process, where $t$ denotes the denoising timestep. This enforces the final generated face images and AU labels to only contain plausible AU activation combinations. Here, the initial co-activation representation $\smash{C_\text{AU}^{(m)}(0)}$ is achieved by CLIP~\cite{radford2021learning} based on the description of every individual AU (examples are provided in Supplementary Material). At each denoising step $t$, the multi-modal joint latent facial representation $\smash{f_\text{joint}^{(m,n)}(t)}$ is updated by a unified condition signal concatenating the textual embedding $\smash{C_\text{T}^{(m)}}$, the identity embedding $\smash{C_\text{ID}^{(n)}}$ extracted by pretrained identity encoder, and the AU co-activation representation $\smash{C_\text{AU}^{(m)}(t-1)}$ encoded from the previous denoising step using the Conditional Label Decoder (CLD) (see Sec.~\ref{sec:DIG}) via cross-attention operation:
\begin{equation}
f_\text{joint}^{(m,n)}(t) 
\sim 
p\!\left(f_\text{joint}^{(m,n)}(t-1)
\mid 
\epsilon(x_t),
C_\text{cond}^{(m,n)}(t-1),
t
\right)
\end{equation}
where 
\(
C_\text{cond}^{(m,n)}(t-1)
=[C_\text{T}^{(m)},\,C_\text{ID}^{(n)},\,C_\text{AU}^{(m)}(t-1)].
\)
This process gradually refines the multi-modal joint latent facial representation to properly fuse face identity, AU textual description, and their co-activation constraints. The refined $\smash{f_\text{joint}^{(m,n)}(t)}$ is, in turn, employed to update $\smash{C_\text{AU}^{(m)}(t-1)}$ as $\smash{C_\text{AU}^{(m)}(t)}$ within this denoising step $t$ as:
\begin{equation}
C_\text{AU}^{(m)}(t) \sim p(C_\text{AU}^{(m)}(t-1) | f_\text{joint}^{(m,n}(t)), \quad \text{s.t.} \quad C_\text{AU}^{(m)} \perp C_\text{ID}^{(n)}
\end{equation}
This iterative feedback allows $\smash{C_\text{AU}^{(m,t)}}$ to evolve from static textual priors into adaptive, context-aware cues, establishing a closed mutual conditioning loop that ensures coherent semantic alignment and facial structural consistency. To avoid circular dependencies, each denoising step computes the multi-modal joint latent facial representation $\smash{f_\text{joint}^{(m,n)}(t)}$ using the \textbf{detached} AU co-activation representation $\smash{C_\text{AU}^{(m)}(t-1)}$ refined from the previous denoising step. 
In addition, during this process, the expression cues in face identity images are mitigated through explicit structural constraints achieved by the Identity Decoupling Modules (IDMs) (see Sec.~\ref{sec:DIG}).

\subsection{Diffusion-based Image-label Joint Generator} 
\label{sec:DIG}

\noindent To jointly synthesize identity-varied facial images and identity-agnostic AU labels, our DIG module decodes the joint representation $\smash{f_\text{joint}^{(m,n)}}$ via dual-branchs: one for generating facial images with diverse identities and the other for producing AU labels independent of identity cues.

\noindent \textbf{Multi-identity facial expression images generation:} This branch uses the upsampling path of the denoising UNet to generate diverse identity-rich facial images. Specifically, we initialize the image latent $z_t$ from the joint latent facial representation $\smash{f_\text{joint}^{(m,n)}}$, and iteratively denoising via \textit{Cross-Attention} conditioned on $\smash{ {C_{\text{AU}}, C_{\text{T}}, C_{\text{ID}}}}$. At each step, $z_{t-1}$ is sampled as:
\begin{equation}
    z_{t-1}\sim \mathcal{N}\left( \mu_\theta(z_t, [C_{\text{AU}}, C_{\text{T}}, C_{\text{ID}}]),  \sigma_t^2 \mathbf{I} \right)
\end{equation}
Here, the $C_{\text{ID}}$ controls identity-specific facial features, while $\smash{C_{\text{AU}}}$ injects the AU co-activation patterns. $ \mu_\theta $ predicts the conditional mean, and $ \sigma_t^2 $ controls the noise level. The $ z_0 $ is decoded into the facial image $ x_{\text{gen}} $.

\noindent \textbf{Identity-agnostic AU label generation:} As shown in~\Cref{fig:CLD}$(a)$, the label generation branch uses a Transformer-based \textit{\textbf{Conditional Label Decoder (CLD)}} to decode AU labels from image latents $\smash{z^{(l)}}$ obtained at different layers of the denoising upsampling path. 
Since the influence of identity-related facial characteristics may degrade accurate AU label generation, we introduce a set of \textit{\textbf{Identity Decoupling Modules (IDMs)}} to suppress identity-related but AU-unrelated features from $ \smash{z^{(l)}}$ via two complementary mechanisms (shown in~\Cref{fig:CLD}$(b)$) as:
\begin{itemize}
    \item \textit{Modulation Mask:} We define a soft attention mask as the complement to the sigmoid similarity between the projected identity and key embeddings, serving as an attention gating that downweights identity-aligned features.
    \begin{equation}
        M_\text{ID}^{(l)} = 1 - \text{Sigmoid}\left( \text{MLP}_{MM}^{(l)}(C_\text{ID}) \cdot K^{(l)\top} \right)
    \end{equation}
    
    \item \textit{Residual Filtering:} A residual filter eliminates identity-specific cues by subtracting the latent’s projection onto normalized vectors $\smash{ \hat{C}_{\text{ID}}^{(l)} }$, derived from $ \smash{C_\text{ID}} $ using lightweight MLPs for dimension alignment:
    \begin{equation}
    \begin{aligned}
    z_\text{res}^{(l)} &= z^{(l)} - \langle z^{(l)}, \hat{C}_\text{ID}^{(l)} \rangle \cdot \hat{C}_\text{ID}^{(l)} \\
    \hat{C}_\text{ID}^{(l)} &= \text{Norm}(\text{MLP}_{\text{RF}}^{(l)}(C_\text{ID}))
    \end{aligned}
\end{equation}

\end{itemize}
Together, they decouple identity signals both at the attention level and content level, ensuring more robust AU structure learning. Although $\smash{z^{(l)}}$ has deviated from the joint representation, the consistent guidance of $C_\text{ID}$ throughout generation makes the suppression still effective. The purified features are computed following standard attention:
{\small
\begin{equation}
\begin{aligned}
z_{\perp C_\text{ID}}^{(l)} = \text{LN} \Big(
\text{Softmax} \Big( \frac{Q^{(l)} K^{(l)\top}}{\sqrt{d_k}} 
\odot M_\text{ID}^{(l)} \Big) V^{(l)} + z_\text{res}^{(l)} \Big)
\end{aligned}
\end{equation}
}

The identity-suppressed features $\smash{z_{\perp C_\text{ID}}^{(l)}}$ are then fused with $N_q$ (number of AUs) learnable AU queries, initialized using CLIP encoded AU descriptions, to enable category-aware alignment. However, prompt semantics may be diluted by dominant visual signals during generation~\cite{baltruvsaitis2018multimodal}. To mitigate this, we introduce a \textit{\textbf{Language-Guided Feature Enhancer (LGFE)}}, interleaved within the Transformer decoder. It applies \textit{Self-Attention} over token-wise prompt embeddings to extract global semantic context, followed by \textit{Cross-Attention} to inject it into AU queries, as shown in~\Cref{fig:CLD}$(c)$. The refined AU queries are passed to a dual-branch prediction head comprising two independent MLPs to estimate AU occurrence $\smash{v_\text{gen}^{(occ)}}$ and intensity $\smash{v_\text{gen}^{(int)}}$, capturing both the activation status and variation level. Through attention-based interactions, AU queries embed co-activation patterns and structural dependencies among facial muscles. A dynamic cross-timestep Exponential Moving Average (EMA) over final-layer AU queries distills these semantics into $\smash{C_\text{AU}^{(m)}}$, serves as a condition in subsequent representation learning.

\begin{table*}[t]
    \centering
    \fontsize{9pt}{9pt}\selectfont
    \setlength{\tabcolsep}{2.5pt} 
    \begin{tabular}{llccccccccc}
        \toprule 
        \textbf{Metrics} &\textbf{Methods} & \textbf{AU1} & \textbf{AU2} & \textbf{AU4} & \textbf{AU6} & \textbf{AU9} & \textbf{AU12} & \textbf{AU25} & \textbf{AU26} & \textbf{Avg.} \\
        \midrule
        \multicolumn{11}{l}{\textit{Metrics for AU Occurence Detection}} \\
        \midrule
        \multirow{5}{*}{\textbf{F1} $\uparrow$}
        &\makecell[l]{FMAE (ViT-B)} & 51.1 / \textbf{54.6} & 55.1 / \textbf{57.4} & 78.6 / \textbf{72.9} & 56.5 / \textbf{60.5} & 44.3 / \textbf{43.2} & 78.9 / \textbf{77.9} & 87.8 / \textbf{86.6} & 55.9 / \textbf{59.8} & 63.5 / \textbf{64.1} \\
        &\makecell[l]{FMAE (ViT-L)} & 47.1 / \textbf{49.5} & 45.7 / \textbf{48.9} & 71.4 / \textbf{74.1} & 50.1 / \textbf{54.4} & 49.5 / \textbf{54.6} & 78.2 / \textbf{82.0} & 80.6 / \textbf{88.5} & 58.9 / \textbf{59.7} & 60.2 / \textbf{63.9} \\
        &\makecell[l]{FMAE (ViT-H)} & 44.5 / \textbf{50.6} & 32.5 / \textbf{36.9} & 47.6 / \textbf{75.9} & 69.7 / \textbf{63.3} & 46.9 / \textbf{47.3} & 77.8 / \textbf{81.2} & 80.4 / \textbf{91.5} & 58.2 / \textbf{59.9} & 57.2 / \textbf{63.3} \\
        &\makecell[l]{GraphAU (Res50)}  & 54.6 / \textbf{59.0} & 47.1 / \textbf{41.6} & 72.9 / \textbf{72.5} & 54.0 / \textbf{63.5} & 55.7 / \textbf{53.9} & 76.7 / \textbf{83.4} & 91.1 / \textbf{82.2} & 53.0 / \textbf{71.8} & 63.1 / \textbf{66.0} \\
        &\makecell[l]{GraphAU (Swin-B)} & 52.5 / \textbf{61.9} & 45.7 / \textbf{47.6} & 76.1 / \textbf{74.2} & 51.8 / \textbf{66.4} & 46.5 / \textbf{36.7} & 76.1 / \textbf{82.8} & 92.9 / \textbf{86.1} & 57.6 / \textbf{73.7} & 62.4 / \textbf{66.2} \\
        \midrule
        \multicolumn{11}{l}{\textit{Metrics for AU Intensity Estimation}} \\
        \midrule
        \multirow{3}{*}{\textbf{ICC} $\uparrow$}
        &\makecell[l]{KJRE}  & .27 / \textbf{.30} & .35 / \textbf{.32} & .25 / \textbf{.22} & .51 / \textbf{.48} & .31 / \textbf{.35} & .67 / \textbf{.70} & .74 / \textbf{.76} & .25 / \textbf{.22} & .42 / \textbf{.41} \\
        &\makecell[l]{SCC-heatmap} & .73 / \textbf{.77} & .44 / \textbf{.42} & .74 / \textbf{.67}  & .27 / \textbf{.30} & .51 / \textbf{.55} & .71 / \textbf{.74} & .94 / \textbf{.92} & .78 / \textbf{.82} & .64 / \textbf{.65} \\
        &\makecell[l]{MAE-face}  & .73 / \textbf{.78} & .66 / \textbf{.70} & .76 / \textbf{.72} & .65 / \textbf{.68} & .60 / \textbf{.56} & .87 / \textbf{.85} & .95 / \textbf{.96} & .75 / \textbf{.69} & .75 / \textbf{.74} \\
      
        \midrule
        \multirow{3}{*}{\textbf{MAE} $\downarrow$}
        &\makecell[l]{KJRE}  & 1.02 / \textbf{.95} & .92 / \textbf{.97} & 1.86 / \textbf{1.62} & .79 / \textbf{.85} & .87 / \textbf{.82} & .77 / \textbf{.90} & .96 / \textbf{.92} & .94 / \textbf{.98} & 1.02 / \textbf{1.00} \\
        &\makecell[l]{SCC-heatmap}  & .16 / \textbf{.20} & .16 / \textbf{18} & .27 / \textbf{.24} & .25 / \textbf{.28} & .13 / \textbf{.18} & .32 / \textbf{.30} & .30 / \textbf{.27} & .32 / \textbf{.35} & .24 / \textbf{.25} \\
        &\makecell[l]{MAE-face}  & .11 / \textbf{.08} & .09 / \textbf{.07} & .28 / \textbf{.32} & .25 / \textbf{.24} & .13 / \textbf{.17} & .21 / \textbf{.25} & .18 / \textbf{.16} & .22 / \textbf{.28} & .18 / \textbf{.19} \\
        \bottomrule
        
    \end{tabular}
    \caption{
        \fontsize{10pt}{10pt}\selectfont AU label consistency is assessed using F1 scores over 8 shared AUs with multiple state-of-the-art detectors. Left and bolded right values indicate results on real data from DISFA and generated samples, respectively. Intensity accuracy is measured by the Intra-class Correlation Coefficient (ICC) and Mean Absolute Error (MAE). }
    \label{tab:au_accuracy}
\end{table*}

\subsection{Training Objective}
\label{sec:Training objective}

\noindent \textbf{Self-supervised Text–Label–Image Consistency:} Beyond architectural enhancements, we introduce the triplet self-supervised objectives to reinforce modality alignment: (1) \textit{Text–Image alignment:} We adopt a local directional CLIP loss~\cite{gal2022stylegan} to guide expression semantics. Unlike absolute similarity, which may overfit identity-preserving content, this loss emphasizes whether the semantic shift from the identity image aligns with the prompt-induced expression change; (2) \textit{Text–Label alignment:} To align AU predictions with the expression semantics, we supervise the generated AU intensities $\smash{v_\text{gen}^{(int)}}$ using the predefined AU vector $\smash{v_\text{pre}^{(int)}}$ that was used to guide prompt generation:
\begin{equation}
\mathcal{L}_{\text{text-AU}} = \| v_\text{gen}^{(int)} - v_\text{pre}^{(int)} \|_2^2.
\end{equation}
This encourages the label outputs to remain consistent with the expression semantics encoded in text; and (3) \textit{Image–Label alignment:} Since prompts are constructed from predefined AU annotations in the training process, they may introduce annotation bias into the learning process. We further supervise the predicted AU occurrences and intensities using labels jointly decided by multiple open-sourced state-of-the-art AU detectors, followed by a cross-detector agreement strategy that evaluates prediction consistency across multiple auxiliary detectors, retaining only consensus labels for backpropagation. This mitigates the biases of each single AU detector. To avoid noisy supervision, AU extraction is restricted to images classified as realistic by a pretrained discriminator, ensuring structural reliability For the full formulation and details of cross-detector agreement strategy, please refer to \textit{Appendix A.10}).

\begin{figure}[t]
    \centering
    \includegraphics[width=0.47\textwidth]{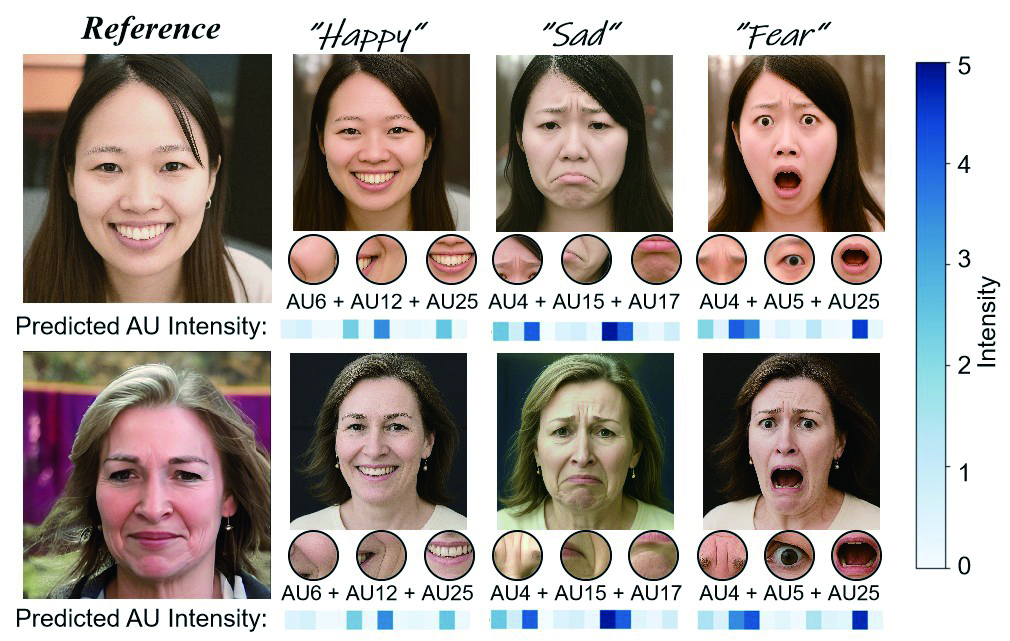} 
    \caption{MAUGen-generated expressions from prompts across identities, with predicted AU intensities. }
    \label{fig:identity}
\end{figure}

\noindent \textbf{Masked Denoising Loss for Facial Regions:} The DDPM-style noise prediction loss~\cite{ho2020denoising} is applied in a spatially selective manner during latent diffusion, with a predefined facial mask $x_\text{mask}$ guiding the optimization toward semantically relevant regions:
\begin{equation}
\mathcal{L}_{\text{mask}} =  \| x_\text{mask} \odot \epsilon - x_\text{mask} \odot \epsilon_\theta(x_{t}, C_{\text{AU}}, C_{\text{T}}, C_{\text{ID}}, t) \|_2^2
\end{equation}

\noindent \textbf{Total loss and Adaptive Loss Weighting:} Our total loss comprises a triplet loss, a masked denoising loss and an identity loss adopted from ArcFace~\cite{deng2019arcface}. While manual weighting of multi-objective losses often requires extensive tuning and can lead to unstable/suboptimal training due to scale mismatch across loss terms, we apply an adaptive weighting based on the relative gradient norms of each loss with respect to the final convolutional layer of the MRL module with numerical stability factor $\delta=10^{-6}$:
\begin{equation}
    \mathcal{L}_{\text{total}} = \sum_{i} \lambda_i \mathcal{L}_i  \,,\quad \lambda_i = \frac{\|\nabla_{\theta} \mathcal{L}_i\|}{\sum_{j} \|\nabla_{\theta} \mathcal{L}_j\|+\delta}
\end{equation}

\begin{table}[t]
    \centering
    \fontsize{9pt}{9pt}\selectfont
    \setlength{\tabcolsep}{1mm}
    \begin{tabular}{@{}ccccccccccc@{}}
        \toprule
        \textbf{CLD} & \textbf{ID} & \textbf{MM} & \textbf{RF}
        &\textbf{AC} &\textbf{LI} & \textbf{LG} & \textbf{$\mathcal{L}_{IL}$} & \textbf{F1 $\uparrow$} & \textbf{FID $\downarrow$} & \textbf{CLIP $\uparrow$} \\ 
        \midrule
        & & & & & & & 
        & --   & 6.2042 & 0.2661 \\ 
        \checkmark & & & & & & & 
        & 92.6 & 6.0823 & 0.2425 \\ 
        \checkmark &\checkmark & & & & & &
        & 88.5 & 6.0795 & 0.2237 \\ 
        \checkmark &\checkmark &\checkmark & & & & &
        & 92.7 & 6.0231 & 0.2681 \\  
        \checkmark &\checkmark &\checkmark &\checkmark & & & &
        & 93.6 & 5.9120 & 0.2769 \\ 
        \checkmark &\checkmark &\checkmark & \checkmark &\checkmark & & &
        & 94.2 & 5.8429 & 0.2817 \\ 
        \checkmark &\checkmark &\checkmark & \checkmark & \checkmark & &\checkmark  & & 95.4 & 5.8170 & 0.2838 \\
        \checkmark &\checkmark &\checkmark & \checkmark & \checkmark &\checkmark & &
        & 95.7 & 5.6185 & 0.2889 \\
         \checkmark &\checkmark &\checkmark & \checkmark & \checkmark &\checkmark  &\checkmark  &
        & 96.2 & 5.5772 & 0.2895 \\
        \checkmark &\checkmark &\checkmark & \checkmark & \checkmark &\checkmark  &\checkmark  & \checkmark
        & \textbf{96.7} & \textbf{5.5163} & \textbf{0.2949} \\
        \bottomrule
    \end{tabular}
    \caption{\fontsize{10pt}{10pt}\selectfont Ablation Study. ``MM'' refers to modulation mask, ``RF'' to residual filtering, ``MC''  to mutual conditioning, ``LI'' to language initialization of AU queries, ``LG'' to LGFE, and ``$\mathcal{L}_{IL}$'' to the image–label alignment loss.
}
    \label{tab:alb_cld}
\end{table}

\section{Experiments}

\subsection{Experimental Settings}

\textbf{Implementation details:} Our MAUGen builds on Stable Diffusion~\cite{rombach2022high,Rombach_2022_CVPR}, with the identity encoder adapted from F2D~\cite{shiohara2024face2diffusion}. 
The main training is conducted on DISFA, using both its 12 AU occurrence and intensity (ranging from 0 to 5) labels. To avoid any potential information leakage, we follow the three-fold data splitting protocol, ensuring that the training and test subsets are completely disjoint.
Our training is achieved using the Adam optimizer on a single NVIDIA L20 GPU. More details on training refer to \textit{Appendix A.2}.

\noindent \textbf{Evaluation Metrics:} We evaluate the quality of generated face images using Fréchet Inception Distance (FID)~\cite{heusel2017gans} and CLIP Score~\cite{radford2021learning}. The consistency of the generated AU labels and face images is measured using three variants of \texttt{FMAE}~\cite{ning2024representation} and two variants of \texttt{GraphAU}~\cite{luo2022learning} AU recognition models, i.e., they predict AU occurrences from the generated face images, which are then compared with their occurrences described in the textual prompts. Mean F1 score is computed and averaged across the comparison results of five variants. This multiple variant strategy utilises the complementary strengths of different architectures, as their recognition performance varies across individual AUs. Meanwhile, \texttt{KJRE}~\cite{KJRE}, \texttt{SCC-heatmap}~\cite{fan2020facialactionunitintensity}, and \texttt{MAE-Face}~\cite{ma2022facialactionunitdetection} are used to predict AU intensities, where intra-class correlation coefficient (ICC) and mean absolute error (MAE) are employed to compare them with intensity status described in the textual prompt.

\begin{table}[t]
    \centering
    \renewcommand{\arraystretch}{1.1}
    \setlength{\tabcolsep}{1.3mm} 
    
    \begin{tabular}{l|ccc}
        \toprule
        \makecell[l]{\textbf{Models}} & \textbf{Real} & \textbf{Syn.} & \textbf{Real+Syn.}  \\ \hline
        \makecell[l]{FMAE (ViT-B)} & 63.5 & 67.1 & \textbf{70.9} \\ \hline
        \makecell[l]{FMAE (ViT-L) } & 63.9 & 66.8 & \textbf{70.4} \\ \hline
        \makecell[l]{FMAE (ViT-H) } & 61.7 & 66.6 & \textbf{67.9} \\ \hline
        \makecell[l]{GraphAU (Res50)} & 63.1 & 63.5 & \textbf{64.6} \\ \hline
        \makecell[l]{GraphAU (Swin-B)} & 62.4 & 63.5 & \textbf{64.1} \\ 
        \bottomrule
    \end{tabular}
    \caption{\fontsize{10pt}{10pt} F1 scores of state-of-the-art AU recognition models trained with real(R) / synthesis(S) datasets.}
    \label{tab:detection_models}
\end{table}

\subsection{Qualitative Evaluation}
\label{sec:qualitative}

\Cref{fig:comparison} presents side-by-side qualitative results across different prompts. It is clear that our MAUGen consistently generates photorealistic face images that show better alignment with the target expressions, surpassing prior methods in both visual fidelity and semantic accuracy. For example, the “expression "chin raise” in Prompt 3 is accurately rendered only by MAUGen, while other models produce ambiguous outputs with inadequate lower-face lifting. This advantage is further evidenced by complex expressions involving fine-grained AU combinations, such as AU4 (brow lowerer) combined with AU9 (nose wrinkler). As shown in~\Cref{fig:identity}, MAUGen also excels in generating identity-agnostic expressions where the prompts are expanded from emotion keywords into detailed AU-based descriptions. The resulting outputs exhibit high inter-identity consistency in AU activation while maintaining facial identity coherence. More extensive comparisons, including failure cases and interpretative analyses, are provided in \textit{Appendix B.5–B.6}.

\subsection{Quantitative Evaluation}

We compare our MAUGen with other methods that are also utilized DISFA and BP4D datasets for fine-tuning, where identity exemplars are also sampled from them for fair comparison. As shown in~\Cref{tab:quantitative_comparisons}, MAUGen consistently achieved lower FID scores, reflecting enhanced visual realism and structural coherence, as well as competitive CLIP scores with only a slight drop on DISFA likely due to its varied background conditions~\cite{radford2021learning}. To assess AU label quality, we evaluate both occurrence and intensity using multiple state-of-the-art AU detectors not seen during training. As reported in~\Cref{tab:au_accuracy}, while detector-specific variations are observed, the trends largely align with the results testing on real data, further supporting the credibility of the synthesized AU labels. We also conduct controlled experiments to examine label accuracy, identity preservation, and prompt sensitivity. 5,000 identities are sampled under fixed prompts to measure identity similarity 
and label variance, with results detailed in \textit{Appendix B.6–B.10}. In addition, certified FACS experts manually inspect randomly selected samples and confirm that the generated AU patterns exhibit plausible activation dynamics, with an ICC of 0.79 (see \textit{Appendix B.3}), which is closely compared to the manual annotation accuracy on DISFA.

\begin{figure}[t]
    \centering
    \includegraphics[width=0.48\textwidth]{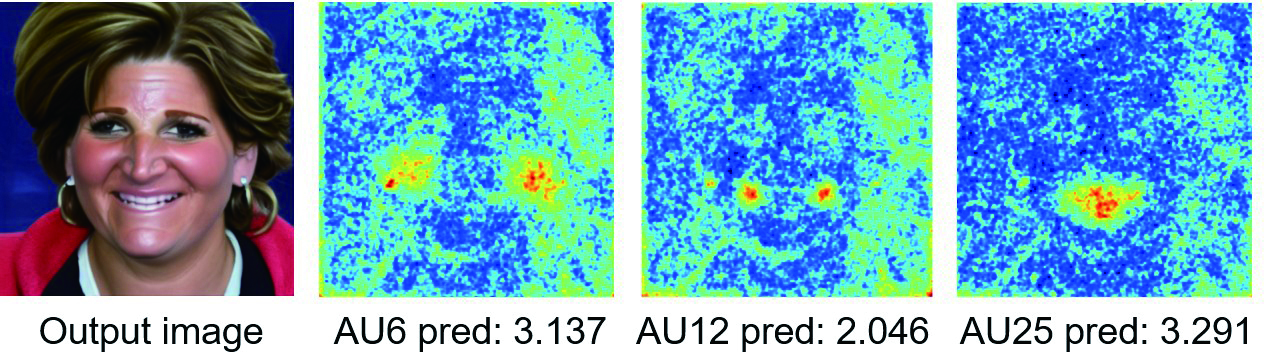} 
    \caption{AU query attention maps focusing on expression-relevant regions with predicted intensities.}
    \label{fig:au_attention}
\end{figure}

\subsection{Ablation Study}
We present the main ablation experiments here, including the analysis of design components, with additional analysis of visual effect of them and adaptive loss in \textit{Appendix B.7}.

\noindent \textbf{Effect of IDM:}
As shown in~\Cref{tab:alb_cld}, introducing identity features without proper decoupling leads to a decrease in F1 and CLIP scores, suggesting that the identity-specific signals are inherent in label generation. Incorporating IDM significantly improves F1 from 88.5 to 93.6, alongside enhancements in CLIP and FID. Note that the F1 score is computed against the predefined AU vectors in prompt construction. We further disentangle the contributions of the modulation mask and residual filtering, both of which yield measurable gains. These results demonstrate that IDM effectively removes identity-specific signals from AU representations, resulting in more stable and identity-agnostic label generation. This effect is also visually confirmed in~\Cref{fig:identity}, where predicted AU labels remain consistent across identities under the same prompt. We further conduct qualitative ablation studies under fixed AU prompts to examine IDM’s influence, showing clearer and more consistent expressions. (visualization results in \textit{Appendix B.7}).

\begin{table}[t]
    \centering
    \fontsize{9pt}{9pt}\selectfont
    \setlength{\tabcolsep}{1.3mm}
    \begin{tabular}{lcccc}
    
    \toprule
    \multirow[b]{2}{*}{\textbf{Models}} & \multicolumn{2}{c}{\textbf{AffectNet}} & \multicolumn{2}{c}{\textbf{Aff-Wild2}} \\

    \cmidrule(lr){2-3} \cmidrule(lr){4-5}
     & DISFA & MIFA & DISFA & MIFA \\
    \midrule
     GraphAU (Res50) & 53.21 & 56.01 & 21.45 & 26.72 \\
     GraphAU (Swin-B) & 61.84 & 65.54 & 21.21 & 21.26 \\
     FMAE (Vit-B) & 56.74 & 53.94 & 38.96 & 40.01 \\
     FMAE (Vit-H) & 70.58
 & 69.49 & 36.39 & 42.12 \\
    \bottomrule
    \end{tabular}
    \caption{Cross-dataset robustness of AU detection models trained on DISFA or MIFA.}
    \label{tab:robustness}
    \end{table}
    
\noindent \textbf{Effect of Mutual Conditioning:}
Building upon IDM, we further evaluate the effectiveness of AU-guided mutual conditioning within the CLD. As shown in~\Cref{tab:alb_cld}, this integration leads to improvements in CLIP scores, indicating a better semantic alignment. As shown in~\Cref{fig:au_attention}, final-layer attention maps reveal focused responses on expression-relevant regions, such as cheeks and lip corners. The observed co-activation across these areas suggests that the model captures structurally coherent AU patterns, validating the effectiveness of AU-guided conditioning.

\begin{figure}[t]
    \centering
    \includegraphics[width=0.47\textwidth]{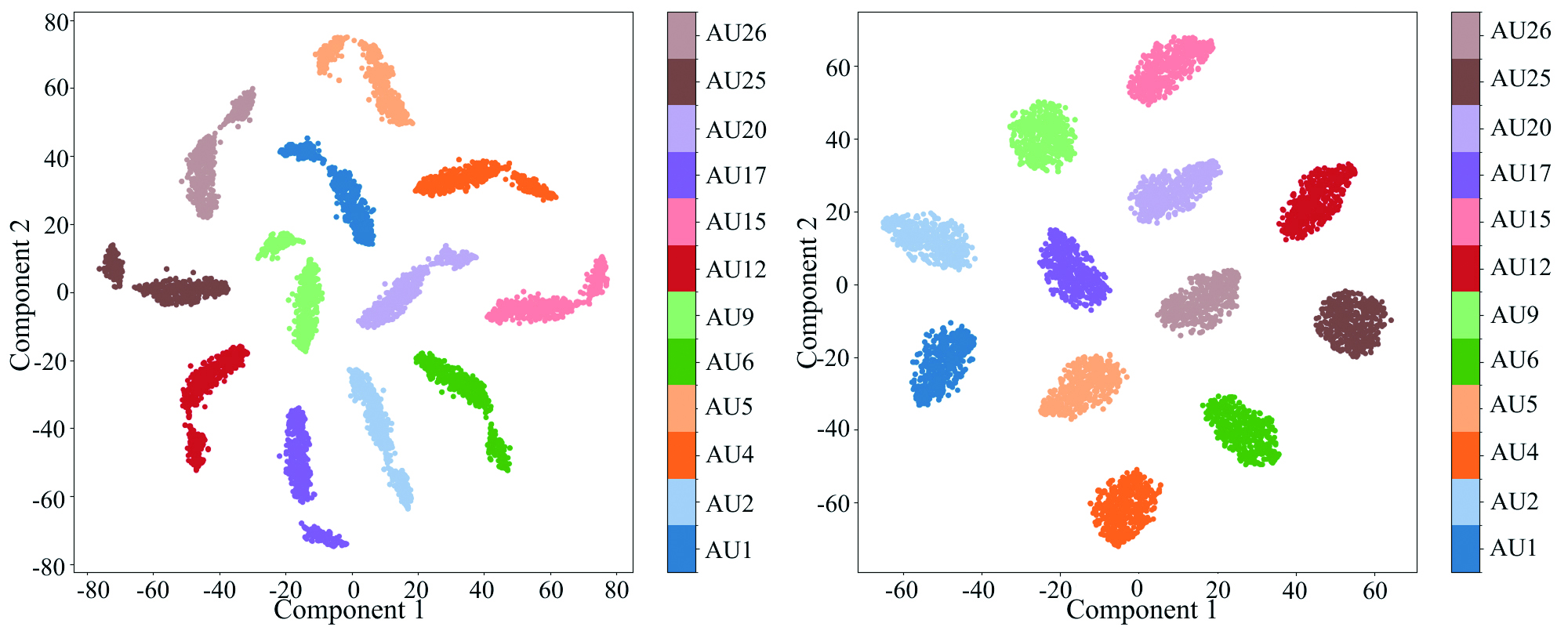} 
    \caption{\textit{t}-SNE of AU queries. \textit{Left}: without language guidance. \textit{Right}: with language guidance. The optimization strategy leads to more compact query clustering. }
    \label{fig:tsne}
\end{figure}

\noindent \textbf{Effect of Language-guided Query Optimization:}
To evaluate the effectiveness of language-guided AU query optimization, we compare a baseline model (without query initialization and LGFE) against the full model. Quantitatively, as reported in~\Cref{tab:alb_cld}, Language-based query initialization improves the score to 95.7, and adding LGFE further raises it to 96.2. As shown in~\Cref{fig:tsne}, the t-SNE visualization of final-layer AU queries reveals that our method yields more compact and well-separated clusters, indicating enhanced discriminability of the query tokens. These improvements in evaluation scores and clustering quality confirm the effectiveness of language-guided initialization and LGFE in enhancing expression controllability.

\noindent \textbf{Effect of Text-Image Alignment Mechanism:}
As shown in~\Cref{tab:alb_cld}, incorporating the pseudo-label loss and cross-detector agreement filtering yields consistent yet marginal improvements, confirming their ability to suppress detector-specific noise and enhance the reliability of semantic supervision.  
These mechanisms encourage the model to align textual cues with visual evidence more faithfully, particularly for subtle or low-intensity AUs.  
However, the overall gains remain moderate, likely due to the inherent domain bias of AU detectors trained on constrained datasets such as DISFA and BP4D, which limits the transferability and diversity of the supervision signals in open-domain generative settings.

\noindent \textbf{Effect of the Synthesized Dataset:}
To assess the utility of synthesized MIFA for AU recognition, we train existing AU detection models on (i) real data from DISFA, (ii) synthetic data, and (iii) a combination of both, and evaluate on real data. As shown in~\Cref{tab:detection_models}, models trained on the combined set consistently outperform those trained on real or synthetic data alone. These gains indicate that our synthesized samples effectively fill coverage gaps in the real dataset, particularly by introducing more diverse AU co-occurrence patterns and better-balanced distributions across both AUs and subjects (see further analysis in \textit{Appendix B.6}).

\noindent \textbf{Robustness of the MIFA Dataset:} The robustness of the proposed MIFA dataset was examined through a cross-dataset evaluation, which assesses its capacity to support stable performance across diverse facial expression domains. Two architectures, GraphAU with ResNet-50 and Swin-B~\cite{luo2022learning}, were each trained separately on the DISFA and MIFA, and both evaluated on the AffectNet and Aff-Wild2 test sets. 500 images were selected and manually annotated for each target dataset. The results listed in \Cref{tab:robustness} show that models trained on MIFA consistently outperform those trained on DISFA on both AffectNet and Aff-Wild2, confirming MIFA’s superior in supporting robust cross-domain affective analysis.

\section{Conclusion}
\label{sec:con}

This work presents the first unified framework for multi-modal facial expression data generation, capable of simultaneously synthesizing photorealistic facial images with preserved identity, identity-agnostic Action Unit (AU) labels, and descriptive textual annotations within a single generative paradigm. The proposed MAUGen framework demonstrates strong cross-modal alignment and expression diversity, leading to high-quality synthetic data that significantly boosts AU recognition performance across multiple benchmarks and model architectures. To facilitate further research, we release MIFA, a standardized and well-curated multi-modal AU dataset constructed using this framework. Overall, our approach offers a scalable and extensible solution for facial expression understanding, with broad potential applications in video synthesis and fine-grained AU modeling. 

\section*{Acknowledgements}
This work was supported by the Natural Science Foundation of Shaanxi Province (Grant No. 2024JC-YBMS-569) and the Key Research and Development Program of Ningbo City (Grant No. 2023Z130)

\bibliography{aaai2026}
\end{document}